\documentclass[11pt,a4paper]{article}
\usepackage[hyperref]{emnlp2018}
\usepackage{times}
\usepackage{latexsym}

\usepackage{url}

\aclfinalcopy % Uncomment this line for the final submission

\title{Neural DrugNet}

\author{Nishant Nikhil \\
  IIT Kharagpur \\
  Kharagpur India\\
  {\tt nishantnikhil@iitkgp.ac.in} \\\And
  Shivansh Mundra \\
  IIT Kharagpur \\
  Kharagpur India\\
  {\tt coolshivansh8@iitkgp.ac.in} \\}

\date{}

\begin{document}
\maketitle
\begin{abstract}
In this paper, we describe the system submitted for the shared task on Social Media Mining for Health Applications by the team Light. Previous works demonstrate that LSTMs have achieved remarkable performance in natural language processing tasks. We deploy an ensemble of two LSTM models. The first one is a pretrained language model appended with a classifier and takes words as input, while the second one is a LSTM model with an attention unit over it which takes character tri-gram as input. We call the ensemble of these two models: Neural-DrugNet. Our system ranks 2nd in the second shared task: Automatic classification of posts describing medication intake.

\end{abstract}

\section{Introduction}

In recent years, there has been a rapid growth in the usage of social media. People post their day-to-day happenings on regular basis. ~\citet{smm4h} propose four tasks for detecting drug names, classifying medication intake, classifying adverse drug reaction and detecting vaccination behavior from tweets. We participated in the Task2 and Task4. 

The major contribution of the work can be summarized as a neural network based on ensemble of two LSTM models which we call Neural DrugNet. We discuss our model in section 2. Section 3 contains the details about the experiments and pre-processing. In Section 4, we discuss the results and propose future works.

\section{Model}
Detection of drug-intake depends highly on:
\begin{itemize} 
\item Whether the sentence conveys an intake.
\item Whether a drug is mentioned in the sentence.
\end{itemize}
Long Short-Term Memory networks~\cite{Hochreiter:1997:LSM:1246443.1246450} have been found efficient in tasks which need to learn structure of a sequential data. To learn a model which can value the first condition, we use LSTM based neural networks. Our first model is inspired from ~\cite{DBLP:journals/corr/abs-1801-06146}, an LSTM model whose encoder is taken from a language model pre-trained on Wikipedia texts and fine-tuned on the tweets. After which a dense layer is used to classify into the different categories.
And to also take into account the mentioning of a drug, which has not been there in the training data, we exploit the word structure of drug-names. Most of the drug-names have the same suffix. Example: melatonin, oxytocin and metformin have the suffix '-in'. We use a LSTM based model trained on the trigrams to learn that. 
Then, we take an ensemble of these two models. We give equal importance to both the models. That is, the prediction probability from Neural-DrugNet is the mean of the prediction probabilities from the two LSTM models. The predicted class is the one having the maximum prediction probability.

The training for the pre-trained language based LSTM model follows the guidelines given in the original paper ~\cite{DBLP:journals/corr/abs-1801-06146}. They use discriminative fine-tuning, slanted triangular learning rates and gradual unfreezing of layers. For the character n-gram based LSTM model, as no fine-tuning is required, we train the model end-to-end. 

\section{Experiments}
% The training datasets tasks are constructed by 15,999 Tweeter
% tweets.   The  testing  set  includes  5000 .  These  are
% collected and manually annotated by the organizers.  Most of the datum has an tweetid, username and is classified into
% one of the three classes: Personal Medication Intake (1), Possible Medication intake (2), Non-Intake (3)
% The distribution of the classes in the training and test dataset  are shown in Table 1.   As
% the user id and user name is not important for prediction,  we decided to drop them off.

% \begin{tabular}{ |p{2cm}||p{2cm}|p{2cm}|  }
%  \hline
%  \multicolumn{3}{|c|}{Table 1} \\
%  \hline
%  Class & Train &Test\\
%  \hline
%  Personal Medication Intake   & 3,198    &951\\
 
%  Possible Medication Intake&   5,146  & 1,593\\
 
%  No intake & 7,135 & 2,456\\

%  \hline
% \end{tabular}
% Class distribution in train and test sets
The data collection methods used to compile the dataset for the shared tasks are described in ~\citet{smm4h}.

\subsection{Preprocessing}
Before feeding the tweets to Neural-DrugNet, we use the same preprocessing scheme discussed in ~\citet{aggression}. Then for the character trigram based model, we add an special character '\$' as a delimiter for the word. That is, the character trigrams of 'ram' would be: '\$ra', 'ram' 'am\$'.

\subsection{Results}
We experimented with different type of architectures for both the tasks. Although any classifier like random forest, decision trees or gradient boosting classifier can be used. But due to lack of time, we used only support vector machine with linear kernel as baseline (denoted as LinearSVC).
During development phase, we mistakenly used only accuracy as the metric for task2. The given results are based on a train-validation split of 4:1.
\begin{table}[h]
\begin{tabular}{ |p{3cm}|p{3cm}|  }

  \hline
 System & Accuracy\\
 \hline
  \hline
 LinearSVC & 0.675\\
  \hline
 LSTM with attention (words as input) & 0.703\\
  \hline
 1D-CNN & 0.651 \\
  \hline
 Bi-LSTM with attention (words as input) & 0.714\\
  \hline
 Bi-LSTM with attention (3-grams as input) & 0.709\\
  \hline
 LSTM with encoder from Language Model & 0.754\\
  \hline
 \textbf{Neural DrugNet} & \textbf{0.771}\\
 \hline
\end{tabular}
 \caption{Results on validation data for Task2}
\end{table}

\begin{table}[h]
\begin{tabular}{ |p{3cm}|p{3cm}|  }
 \hline
 System & F1-score\\
 \hline
  \hline
 LinearSVC & 0.751\\
  \hline
 Neural DrugNet & 0.805\\
  \hline
 \textbf{Neural DrugNet with LM fine-tuned on data from task3 also} & \textbf{0.812}\\
 \hline
\end{tabular}
 \caption{Results on validation data for Task4}
\end{table}

The final results on best performing variant on test data for both the tasks are:

\begin{table}[h]
\begin{tabular}{ |p{2cm}|p{2cm}|p{2cm}|  }
 \hline
 Precision & Recall & F1-score\\
 \hline
 0.520 & 0.491 & 0.505 \\
 \hline
 \end{tabular}
 \caption{Results on test data for Task2}
\end{table}

\begin{table}[h]
\begin{tabular}{ |p{1.5cm}|p{1.5cm}|p{1.5cm}|p{1.5cm}|  }
 \hline
 Accuracy & Precision & Recall & F1-score\\
 \hline
 0.857 & 0.824 & 0.897 & 0.859 \\
 \hline
 \end{tabular}
 \caption{Results on test data for Task4}
\end{table}

\section{Conclusion and Future Work}
In this paper, we present Neural-DrugNet for drug intake classification and detecting vaccination behavior. It is an ensemble of two LSTM models. The first one is a pretrained language model appended with a classifier and takes words as input, while the second one is a LSTM model with an attention unit over it which takes character tri-gram as input. It constantly outperforms the vanilla LSTM models and other baselines, which supports our claim that drug-intake classification and vaccination behavior detection rely on both the sentence structure and the character tri-gram based features.
The performance reported in this paper could be further boosted by using a language model pre-trained on tweets rather than the wikipedia texts. Furthermore, the ensemble module can be learned end-to-end by using a dense layer.
% By uncommenting {\small\verb|\aclfinalcopy|} at the top of this 
%  document, it will compile to produce an example of the camera-ready formatting; by leaving it commented out, the document will be anonymized for initial submission.

%   \citep{Gusfield:97} & \verb|\citep| & \verb|\cite| \\
%   \citet{Gusfield:97} & \verb|\citet| & \verb|\newcite| \\
%   \citeyearpar{Gusfield:97} & \verb|\citeyearpar| & \verb|\shortcite| \\

% {\bf Citations}: Citations within the text appear in parentheses
% as~\cite{Gusfield:97} or, if the author's name appears in the text
% itself, as Gusfield~\shortcite{Gusfield:97}.

% We suggest that instead of
% \begin{quote}
%   ``\cite{Gusfield:97} showed that ...''
% \end{quote}
% you use
% \begin{quote}
% ``Gusfield \shortcite{Gusfield:97}   showed that ...''
% \end{quote}

\bibliography{emnlp2018}
\bibliographystyle{acl_natbib_nourl}

\end{document}